%% file: ms.tex
\documentclass{esannV2}

\usepackage[latin1]{inputenc}

\input{commands}

\usepackage{tabularx}

\usepackage{amssymb}
\usepackage{graphicx}

\usepackage{standalone}

\usepackage{tikz}
\usetikzlibrary{positioning}
\usetikzlibrary{arrows}
\usetikzlibrary{fit,calc,positioning,decorations.pathreplacing,matrix}

\usepackage{xifthen}

\usepackage{amsmath}
\usepackage[boxed]{algorithm}
\usepackage[noend]{algpseudocode}

\usepackage{setspace}

\usepackage{url}

\begin{document}

\title{Attention-based Information Fusion using Multi-Encoder-Decoder Recurrent Neural Networks}

\author{Stephan Baier$^1$, Sigurd Spieckermann$^2$ and Volker Tresp$^{1,2}$
%
\vspace{.3cm}\\
%
1- Ludwig Maximilian University\\
Oettingenstr. 67, Munich, Germany
%
\vspace{.1cm}\\
2- Siemens AG, Corporate Technology \\
Otto-Hahn-Ring 6, Munich, Germany\\
}

\maketitle

\begin{abstract}
With the rising number of interconnected devices and sensors, modeling distributed sensor networks is of increasing interest. Recurrent neural networks (RNN) are considered particularly well suited for modeling sensory and streaming data. When predicting future behavior, incorporating information from neighboring sensor stations is often beneficial. We propose a new RNN based architecture for context specific information fusion across multiple spatially distributed sensor stations. Hereby, latent representations of multiple local models, each modeling one sensor station, are jointed and weighted, according to their importance for the prediction. The particular importance is assessed depending on the current context using a separate attention function. We demonstrate the effectiveness of our model on three different real-world sensor network datasets.

\end{abstract}

\section{Introduction}

In this paper we propose a recurrent neural network (RNN) architecture for combining information from multiple data streams in a distributed sensor network. With the rising number of connected devices and sensors, often referred to as the Internet of Things (IoT), modeling sensor networks and multi-agent systems is of increasing interest. We consider sensor networks consisting of multiple stations, where each station can measure multiple features at a single location. We address the task of sequence-to-sequence prediction, although our proposed architecture can easily be generalized to other tasks such as classification, recommendation, or anomaly detection. We build dedicated RNN models for all sensor stations, which are allowed to exchange information among each other to enable exploitation of cross-device correlations. The model, which we refer to as the ``multi-encoder-decoder model'', is an extension of the general encoder-decoder framework, which has become popular in various tasks such as machine translation, image caption generation and automatic speech recognition \cite{seq-to-seq}\cite{BengioAttention}. The idea of using multiple encoders and decoders has also recently been considered in natural language processing \cite{google_multi, bengio_multi, waibel_multi}. We propose an interconnection layer, which joins the latent representations of all encoders using an attention mechanism. Thereby, the attention mechanism, which was originally developed for neural machine translation (see \cite{bengio-attention-translation}\cite{BengioAttention}), is applied in a novel context which could also be useful for further sensor fusion tasks. We demonstrate the effectiveness of our proposed multi-sequence-to-sequence network on three datasets. The experimental results show that the proposed attention-based multi-encoder-decoder model outperforms competitive linear models and standard RNN architectures.

\section{Representation Fusion Model}

We propose a model which learns representations for multiple senor stations. The representations are fused using an attention mechanism. Finally, decoder models make predictions based on the fused representations. The whole system is completely differentiable and can thus be trained directly end-to-end. Figure \ref{figure1} shows the model schematically. 

\subsection{Multi-Encoder-Decoder Model}
We consider the task of predicting multiple multivariate output sequences given multiple multivariate input sequences. We apply the sequence-to-sequence model \cite{seq-to-seq} to multiple data streams by creating
multiple encoder and decoder functions. The multiple sequence-to-sequence models communicate through an interconnection layer, which acts like a soft-switching circuit between the single models. The input sequences are represented by a three-way tensor $\mathcal{X} \in \setR^{E \times T_\text{enc} \times F_\text{enc}}$, where $E$ denotes the number of encoder devices, $T_\text{enc}$ denotes the encoder sequence length and $F_\text{enc}$ is the number of encoder features. Similarly, the output sequences are represented by a three-way tensor $\tensor{Y} \in \setR^{D \times T_\text{dec} \times F_\text{dec}}$, where $D$ denotes the number of decoder devices, $T_\text{dec}$ denotes the decoder sequence length and $F_\text{dec}$ is the number of decoder features. In the case of multivariate streaming data from a sensor network, the value $\tensor{X}_{i,t,j}$ corresponds to the $j$-th feature measured at the $i$-th sensor station at time $t$. Similarly, the value $\tensor{\widehat{Y}}_{i,t,j}$ corresponds to the prediction of the $j$-th feature at the $i$-th output node at time $t$. If we consider, for example, the task of predicting the features of the next $T_\text{dec}$ values for all stations in a sensor network, then $D$ is the number of stations, $F_\text{dec}$ is the number of features and $T_\text{dec}$ is the time period for which forecasts are performed.

Each input-sensing device is modeled by an encoder function
\begin{equation}
f_{\text{enc},i}(\tensor{X}_{i, :, :}) = e_i \text{, \hspace{0.2cm} with }  i \in \{1,2,...,E\},
\end{equation}
which takes the data measured at the $i$-th sensing device as input and outputs a latent representation $e_i \in \setR^{\text{dim}(e_i)}$. For each output device an interconnection function $f_{\text{con},j}$ combines the representations $\{e_i\}_{i=1}^{E}$ as
\begin{equation}
f_{\text{con},j}(\{e_i\}_{i=1}^E) = c_j \text{, \hspace{0.2cm} with } j \in \{1,2,...,D\}.
\end{equation}
with $c_j \in \setR^{\text{dim}(e_i)}$. Finally, for each output device a decoder function $f_{\text{dec},j}$ models the prediction given the respective combined representation $c_j$ as
\begin{equation} 
f_{\text{dec},j}(c_j) = \tensor{\widehat{Y}}_{j, :, :} \text{, \hspace{0.2cm} with } j \in \{1,2,...,D\}.
\end{equation}
This way information between the different input and output sequences can be exchanged through the interconnection layer.

\begin{figure*}[t]
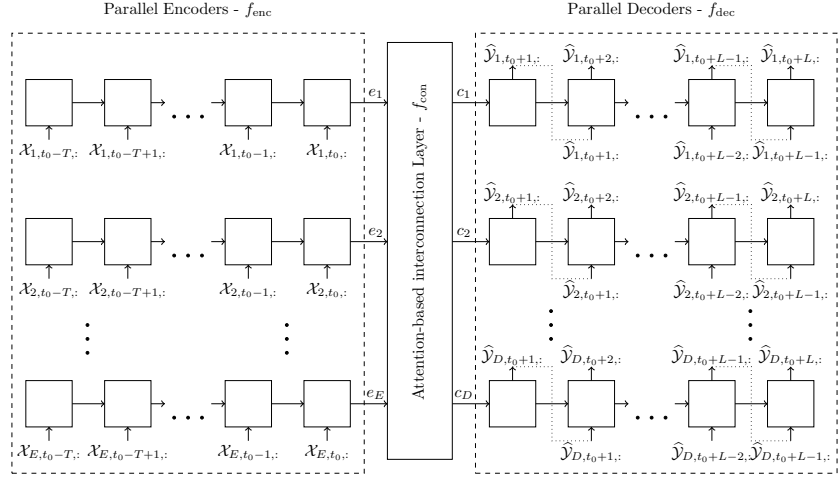

	\centering
	\scalebox{.9}{
		\includestandalone[width=\textwidth]{multi-encoder-decoder3}
	}
	\caption{Unfolded multi-encoder-decoder recurrent neural network for multiple sequence-to-sequence prediction.}
	\label{figure1}
\end{figure*}

Figure~\ref{figure1} shows the architecture of a multi-encoder-decoder recurrent neural network model. For the sequence-to-sequence prediction, we model each encoder and each decoder function with an RNN. Each encoder RNN iterates over the sequence produced by the respective sensing node. Thus, the input of the $i$-th encoder RNN is $x_t = \tensor{X}_{i, t, :}$. We define the last hidden state of the $i$-th encoder RNN to be the encoder output $e_i$. For each decoder RNN a combined representation is computed by the respective interconnection function, which is used as initial hidden representation. The decoder output $\tensor{\widehat{Y}}_{i, t-1, :}$ is copied to the input of the $i$-th decoder RNN at time $t$.

\subsection{Spatial Attention Mechanism}

The interconnection layer is implemented using an attention mechanism, where the combination of latent representations is not fixed for every prediction but depends on the current context, which is encoded in the input representations. The attention mechanism assesses the importance of the representations of the encoding devices $e_i$ and computes a weighted sum over the latent vectors
\begin{equation}
c_j = \frac{1}{E}\sum_{i=1}^{E} w_{ji} e_i,
\end{equation} 
where the weights $w_{ij} \in \setR$ are derived from an additional attention function $f_\text{att}$. The attention function is modeled by an additional feed-forward neural network. The outputs of the attention function are normalized through a softmax function, such that
\begin{subequations}
	\begin{align}
	z_{ji} &= f_{\text{att},j}(e_i) \\
	w_{ji} &= \frac{\exp(z_{ji})}{\sum_{k=1}^{E} \exp(z_{jk})}.
	\end{align}
\end{subequations}
Whether attention is put on a representation $e_i$ or not can vary for each prediction, depending on the encoded information in $e_i$. The approach draws inspiration from the attention-based machine translation model \cite{bengio-attention-translation}, however the attention is not used across time but spatially across sensing devices.

Note that this mechanism can deal with a variable amount of input devices, which is especially useful in settings where the number of input-devices is not constant over time, e.g. moving devices where devices appear and disappear over time, or where some input devices do not
send any data, e.g. due to broken sensors.

\subsection{Model Training}
The complete model is trained end-to-end by minimizing the negative log-likelihood of a historical training set $\set{D} = \{(\tensor{X}^{(n)},\tensor{Y}^{(n)})\}_{n=1}^N$ w.r.t.\ the model parameters such that
\begin{equation} 
L = - \sum_{n=1}^{N} \log \: p(\tensor{Y}^{(n)} | \tensor{X}^{(n)}; \Phi),
\end{equation}
where $\Phi$ includes the parameters of all encoders and decoders, as well as the parameters of the feedforward neural network for the attention function. The cost function is minimized using stochastic gradient descent with mini batches.

\section{Experiments}

We evaluate the performance of the multi-encoder-decoder network using sequence-to-sequence prediction in sensor networks on two climatological datasets and a smart grid dataset. We choose the task to be the prediction of future network behavior given a sequence of past measurements. Predictions are made for every sensor station and all features, thus, $E = D$ and $F_\text{enc} = F_\text{dec}$. 

\subsection{Datasets}

We consider a sensor network of environmental sensing stations measuring climatological data on an hourly basis. The dataset consists of 18 stations distributed across Quebec, each measuring air temperature, dew point, relative humidity and wind speed. The second dataset is a sensor network of 15 environmental sensors spread across Alberta measuring the same features. We downloaded 5 years of data between 2010 and 2014 from ASOS\footnote{https://mesonet.agron.iastate.edu/request/download.phtml} and selected stations and features with the least missing values. We extracted sequences of 72 hours as input to the encoders and made predictions for the next 24 hours. The data gathered between 2010 and 2013 was used for training and validation while the data gathered in 2014 was used for testing the models. In the second experiment we predict the load profiles of the next 3 days given the last 21 (3 weeks) load profiles from certain areas. We selected 18 zones with historical load profiles gathered between 2007 and 2014 from the smart grid dataset \cite{gefcom}. As there is only one measurement we chose the input and target features to be the hourly load and performed the forecasts on a daily basis. 

\subsection{Methods}

We compare our model to multiple linear regression, which has shown state-of-the-art performance in the task of energy load forecasting \cite{gefcom}. Further, we compare against regular RNN models. Both, the linear and the RNN models are trained in two different settings: (i) a separate model for each station, i.e. no cross-correlations can be exploited and (ii) a joint model for all stations, i.e. cross-correlations between stations can be exploited. We evaluate on the normalized data to get a baseline mean squared error of 1.0 for predicting the historical mean. Further, we report as a baseline the constant prediction of the last observed value for each measured feature. For all models the optimal size of the hidden state was determined on the validation set. This resulted in a size of 130 hidden neurons for the RNNs modeling single stations and 300 hidden neurons for the RNNs which model all stations jointly. We also tried the extensions gated recurrent units (GRU) and long short-term memory (LSTM), however the prediction results did not significantly improve. In \cite{schmidhuber-forecasting} it has also been found that LSTMs are not particularly well suited for time series forecasting. All experiments where implemented using Theano \cite{theano}.

\subsection{Results}

\begin{table*}		
	\centering
	\begin{tabular}{|l || c | c | c |}
		\hline
		Dataset & Quebec & Alberta & Smart Grid\\
		\hline
		Last observed values & 65.15 & 72.95 & 51.69\\
		Linear regression per station & 42.89 &  41.89 & 33.82 \\
		Linear regression all stations & 35.62 & 34.87  & 31.64 \\
		Regular RNN per station & 38.17  & 34.92 & 31.50 \\
		Regular RNN all stations & 34.77& 34.68 & 29.56 \\
		Multi-enc-dec RNN attention & 32.28 & 32.89 & 28.84 \\
		\hline
	\end{tabular}
	\vspace{0.2cm}
	\caption{Mean squared error results for the climatological and smart grid test sets in percent.}
	\label{tab:results1}
\end{table*}

Table \ref{tab:results1} shows the results for both datasets. On the climatological dataset we can see that both the RNN and linear model perform significantly better when all stations are integrated into one model compared to one dedicated model for each station. This observation indicates strong cross-correlations between the stations. Using individual RNNs per station performs better than the linear regression model per station, and the joint RNN for all stations outperforms the linear model for all stations. Our proposed multi-encoder-decoder model with spatial attention achieves the best result. This indicates that the attention function helps exploiting the non-linear cross correlations in the overall system. For the smart grid dataset the prediction of the load profile of the last day (last observed values) is already a good baseline as the profiles do not change drastically within three days. Also here the linear model with all stations included slightly improves the prediction over the single models and also the RNN model including all stations outperforms the single per-station RNN models. Also on this dataset, the attention-based multi-encoder-decoder model yields better performance than the baseline models.

\section{Conclusion}

We proposed a neural network architecture for modeling distributed sensor networks, which extends the successful encoder-decoder framework. The fusion of hidden representations of multiple encoder networks using an attention mechanism, allows for exploiting cross-correlations across sensor stations. Using end-to-end training, the complete model consisting of the encoders, the interconnection layer with an attention mechanism, and the decoders is trained to predict a sequence of future behavior. In future work our architecture could also easily be extended to different prediction tasks such as classification or anomaly detection.

\begin{footnotesize}

\bibliographystyle{unsrt}
\bibliography{references}

\end{footnotesize}


\end{document}

%% file: commands.tex
\usepackage{bm}
\newcommand{\tensor}[1]{\ensuremath{\mathcal{#1}}}
\newcommand{\set}[1]{\ensuremath{\mathcal{#1}}}

\newcommand{\setR}{\ensuremath{\mathbb{R}}}

%% file: multi-encoder-decoder3.tex
\begin{tikzpicture}
	
	\foreach \row in {0, 3.5,6.5} {  
		\draw [black] (0, \row) rectangle (1, \row + 1);	
		\draw [black]  (1.7, \row) rectangle (2.7, \row + 1);	
			
		\draw [black]  (4.3, \row) rectangle (5.3, \row + 1);
		\draw [black]  (6, \row) rectangle (7, \row + 1);

		\draw [black]  (10, \row) rectangle (11, \row + 1);	
		\draw [black]  (11.7, \row) rectangle (12.7, \row + 1);	
		
		\draw [black]  (14.3, \row) rectangle (15.3, \row + 1);
		\draw [black]  (16, \row) rectangle (17, \row + 1);
		
		\draw[black, fill=black] (3.2, \row + 0.2) circle (0.2ex);
		\draw[black, fill=black] (3.5, \row + 0.2) circle (0.2ex);
		\draw[black, fill=black] (3.8, \row + 0.2) circle (0.2ex);
		
		\draw[black, fill=black] (13.2, \row + 0.2) circle (0.2ex);
		\draw[black, fill=black] (13.5, \row + 0.2) circle (0.2ex);
		\draw[black, fill=black] (13.8, \row + 0.2) circle (0.2ex);	
		
		\draw [semithick,->] (1, \row + 0.5) -- (1.7, \row + 0.5);
		\draw [semithick,->] (5.3, \row + 0.5) -- (6, \row + 0.5);
		
		\draw [semithick,->] (11, \row + 0.5) -- (11.7, \row + 0.5);
		\draw [semithick,->] (15.3, \row + 0.5) -- (16, \row + 0.5);
		
		\draw [semithick,->] (2.7, \row + 0.5) -- (3, \row + 0.5);
		\draw [semithick,->] (4, \row + 0.5) -- (4.3, \row + 0.5);	
		
		\draw [semithick,->] (12.7, \row + 0.5) -- (13, \row + 0.5);
		\draw [semithick,->] (14, \row + 0.5) -- (14.3, \row + 0.5);	
		
		\draw [semithick, ->] (0.5, \row - 0.3) -- (0.5, \row);
		\draw [semithick, ->] (2.2, \row - 0.3) -- (2.2, \row);
		\draw [semithick, ->] (4.8, \row - 0.3) -- (4.8, \row);
		\draw [semithick, ->] (6.5, \row - 0.3) -- (6.5, \row);
		
		\draw [semithick, ->] (12.2, \row - 0.3) -- (12.2, \row);
		\draw [semithick, ->] (14.8, \row - 0.3) -- (14.8, \row);
		\draw [semithick, ->] (16.5, \row - 0.3) -- (16.5, \row);
		
		\draw [semithick, ->] (10.5, \row + 1) -- (10.5, \row + 1.3);
		\draw [semithick, ->] (12.2, \row + 1) -- (12.2, \row + 1.3);
		\draw [semithick, ->] (14.8, \row + 1) -- (14.8, \row + 1.3);
		\draw [semithick, ->] (16.5, \row + 1) -- (16.5, \row + 1.3);
		
		\draw [dotted] (10.5, \row + 1.3) to (11.35, \row + 1.3) to (11.35, \row - 0.3) to (12.2, \row - 0.3);
		\draw [dotted] (14.8, \row + 1.3) to (15.65, \row + 1.3) to (15.65, \row - 0.3) to (16.5, \row - 0.3);
	
	
		\draw [semithick, ->] (9.2, \row + 0.5) -- (10, \row + 0.5);

		\ifthenelse{\equal{\row}{0}}{
			
			\node at (7.55, \row +0.7) {$e_E$};
			
			\node at (9.45, \row +0.7) {$c_D$};
			
			\node at (0.5, \row - 0.55) {$\mathcal{X}_{E, t_0-T, :}$};
			\node at (2.2, \row - 0.55) {$\mathcal{X}_{E, t_0-T+1, :}$};
			
			\node at (4.8, \row - 0.55) {$\mathcal{X}_{E, t_0-1, :}$};
			\node [rectangle] at (6.5, \row - 0.55) {$\mathcal{X}_{E, t_0, :}$};
			
			\node at (12.2, \row - 0.55) {$\mathcal{\widehat{Y}}_{D, t_0+1, :}$};
			
			\node at (14.8, \row - 0.55) {$\mathcal{\widehat{Y}}_{D, t_0+L-2, :}$};
			\node at (16.5, \row - 0.55) {$\mathcal{\widehat{Y}}_{D, t_0+L-1, :}$};

			\node at (10.5, \row + 1.55) {$\mathcal{\widehat{Y}}_{D, t_0+1, :}$};
			\node at (12.2, \row + 1.55) {$\mathcal{\widehat{Y}}_{D, t_0+2, :}$};
			
			\node at (14.8, \row + 1.55) {$\mathcal{\widehat{Y}}_{D, t_0+L-1, :}$};
			\node at (16.5, \row + 1.55) {$\mathcal{\widehat{Y}}_{D, t_0+L, :}$};		
			
			\draw[semithick, ->] (7, \row + 0.5) -- (7.8, 0 + 0.5);
		}
		
		\ifthenelse{\equal{\row}{3.5}}{
			
			\node at (7.55, \row +0.7)  {$e_2$};
			
			\node at (9.45, \row +0.7) {$c_2$};
			
			\node at (0.5, \row - 0.55) {$\mathcal{X}_{2, t_0-T, :}$};
			\node at (2.2, \row - 0.55) {$\mathcal{X}_{2, t_0-T+1, :}$};
			
			\node at (4.8, \row - 0.55) {$\mathcal{X}_{2, t_0-1, :}$};
			\node [rectangle] at (6.5, \row - 0.55) {$\mathcal{X}_{2, t_0, :}$};
			
			\node at (12.2, \row - 0.55) {$\mathcal{\widehat{Y}}_{2, t_0+1, :}$};
			
			\node at (14.8, \row - 0.55) {$\mathcal{\widehat{Y}}_{2, t_0+L-2, :}$};
			\node at (16.5, \row - 0.55) {$\mathcal{\widehat{Y}}_{2, t_0+L-1, :}$};

			\node at (10.5, \row + 1.55) {$\mathcal{\widehat{Y}}_{2, t_0+1, :}$};
			\node at (12.2, \row + 1.55) {$\mathcal{\widehat{Y}}_{2, t_0+2, :}$};
			
			\node at (14.8, \row + 1.55) {$\mathcal{\widehat{Y}}_{2, t_0+L-1, :}$};
			\node at (16.5, \row + 1.55) {$\mathcal{\widehat{Y}}_{2, t_0+L, :}$};		
			
			\draw [semithick, ->] (7, \row + 0.5) -- (7.8, 3.5 + 0.5);
		}
		
		\ifthenelse{\equal{\row}{6.5}}{
			
			\node at (7.55, \row +0.7) {$e_1$};
			
			\node at (9.45, \row +0.7) {$c_1$};
			
			\node at (0.5, \row - 0.55) {$\mathcal{X}_{1, t_0-T, :}$};
			\node at (2.2, \row - 0.55) {$\mathcal{X}_{1, t_0-T+1, :}$};
			
			\node at (4.8, \row - 0.55) {$\mathcal{X}_{1, t_0-1, :}$};
			\node [rectangle] at (6.5, \row - 0.55) {$\mathcal{X}_{1, t_0, :}$};
			
			\node at (12.2, \row - 0.55) {$\mathcal{\widehat{Y}}_{1, t_0+1, :}$};
			
			\node at (14.8, \row - 0.55) {$\mathcal{\widehat{Y}}_{1, t_0+L-2, :}$};
			\node at (16.5, \row - 0.55) {$\mathcal{\widehat{Y}}_{1, t_0+L-1, :}$};		
			
			\node at (10.5, \row + 1.55) {$\mathcal{\widehat{Y}}_{1, t_0+1, :}$};
			\node at (12.2, \row + 1.55) {$\mathcal{\widehat{Y}}_{1, t_0+2, :}$};
			
			\node at (14.8, \row + 1.55) {$\mathcal{\widehat{Y}}_{1, t_0+L-1, :}$};
			\node at (16.5, \row + 1.55) {$\mathcal{\widehat{Y}}_{1, t_0+L, :}$};		
			
			\draw [semithick, ->] (7, \row + 0.5) -- (7.8, 6.5 + 0.5);			
		}		

	}
	
	
	\draw [black, dashed] (-0.3, -1) rectangle (7.3, 8.5);	
	\draw [black, dashed] (9.7, -1) rectangle (17.5, 8.5);	
	
	\draw [black] (7.8, -0.7) rectangle (9.2, 8.3);
	
	\node at (3.5, 9) {Parallel Encoders - $f_\text{enc}$};
	\node[rotate=90] at (8.5, 4) {Attention-based interconnection Layer - $f_\text{con}$};
	\node at (13.5, 9) {Parallel Decoders - $f_\text{dec}$};
	
	\draw[black, fill=black] (1.33, 2.2) circle (0.2ex);
	\draw[black, fill=black] (1.33, 1.9) circle (0.2ex);
	\draw[black, fill=black] (1.33, 1.6) circle (0.2ex);
	
	\draw[black, fill=black] (5.63, 2.2) circle (0.2ex);
	\draw[black, fill=black] (5.63, 1.9) circle (0.2ex);
	\draw[black, fill=black] (5.63, 1.6) circle (0.2ex);
	
	\draw[black, fill=black] (11.33, 2.5) circle (0.2ex);
	\draw[black, fill=black] (11.33, 2.2) circle (0.2ex);
	\draw[black, fill=black] (11.33, 1.9) circle (0.2ex);
	
	\draw[black, fill=black] (15.63, 2.5) circle (0.2ex);
	\draw[black, fill=black] (15.63, 2.2) circle (0.2ex);
	\draw[black, fill=black] (15.63, 1.9) circle (0.2ex);
	
\end{tikzpicture}